\documentclass[11pt,a4paper]{article}

\usepackage{times}
\usepackage{latexsym}
\usepackage{url}
\usepackage[utf8]{inputenc}
\usepackage[T1]{fontenc}
\usepackage{booktabs}
\usepackage{pgf, tikz}
\usetikzlibrary{arrows, automata}
\usepackage{CJKutf8}
\usepackage{amsmath}
\usepackage{graphicx}
\usepackage{multirow}

\usepackage{caption}
\usepackage{natbib}
\usepackage[margin=25mm]{geometry}

\renewcommand*{\thefootnote}{\fnsymbol{footnote}}
\title{R-grams: Unsupervised Learning of Semantic Units} 
\author{Ariel Ekgren\footnote{These authors contributed equally to the work.} \\
RISE \\
\footnotesize\texttt{ariel.ekgren@gmail.se} \\
\and
Amaru Cuba Gyllensten\textsuperscript{$\dagger$} \\
RISE \\
\footnotesize\texttt{amaru.cuba.gyllensten@ri.se} \\
\and
Magnus Sahlgren\\
RISE \\
\footnotesize\texttt{magnus.sahlgren@ri.se}}

\date{}

\begin{document}
\maketitle

\renewcommand*{\thefootnote}{\arabic{footnote}}
\setcounter{footnote}{0}

\begin{abstract} 
This paper investigates data-driven segmentation using Re-Pair or Byte Pair Encoding-techniques. In contrast to previous work which has primarily been focused on subword units for machine translation, we are interested in the general properties of such segments above the word level. We call these segments r-grams, and discuss their properties and the effect they have on the token frequency distribution. The proposed approach is evaluated by demonstrating its viability in embedding techniques, both in monolingual and multilingual test settings. We also provide a number of qualitative examples of the proposed methodology, demonstrating its viability as a language-invariant segmentation procedure. 
\end{abstract}

\section{Introduction}

Natural Language Processing (NLP) requires data to be segmented into units. These units are normally called {\em words}, which in itself is a somewhat vague and controversial concept \citep{haspelmath:2011} that is often operationalized as meaning something like ``white-space (and punctuation) delimited string of characters''. Of course, some languages do not use white-space delimiters, such as Chinese and Thai, which have context-dependent notions of what constitute words without special symbols dedicated to segmentation. As an example, the sequence \begin{CJK*}{UTF8}{gbsn}我喜欢新西兰花\end{CJK*} can be segmented (correctly) in two different ways \citep{Badino2004ChineseTW}:
\begin{center}
\begin{CJK*}{UTF8}{gbsn}我 / 喜欢 / 新 / 西兰花 \end{CJK*}

I like fresh broccoli 

\begin{CJK*}{UTF8}{gbsn}我 / 喜欢 / 新西兰 / 花 \end{CJK*}

I like New Zealand flowers
\end{center}

Even for white-space segmenting languages, it is seldom as simple as merely using white-space delimited strings of characters as atomic units. As one example, morphologically sparse languages such as English rely to a large extent on word order to encode grammar, which means that such languages often form lexical multi-word units, which by all accounts function as atomic units on the same level as white-space delimited words. As an example, ``white house'' and ``rock and roll'' are both distinct semantic concepts that it would be beneficial to include as atomic units in an NLP application.

Of course, atomic units of language can also exist {\em below} the level of white-space delimited strings of characters. In linguistics, {\em morphemes} are defined as the atomic units of language. For synthetic languages such as Turkish, Finnish, or Greenlandic, where grammatical relations are encoded by morphology rather than word order, there can be a possibly large number of morphemes within one single white-space delimited string of characters. The canonical example in this case tends to be Western Greenlandic, which is a polysynthetic language that produces notoriously long white-space delimited string of characters. As an example, the string ``tusaanngitsuusaartuaannarsinnaanngivipputit'' consists of 9 different morphemes (``hear''|neg.|intrans.participle|``pretend''|``all the time''|``can''|neg.|``really''|2nd.sng.indicat.) and means ``you simply cannot pretend not to be hearing all the time''. One white-space delimited string of characters in Western Greenlandic, eleven in English. 

Similarly, compounding languages such as Swedish can form productive compounds, where a potentially large number of words (and morphemes) are compounded into one single white-space delimited string of characters. As an example, the string ``forskningsinformationsförsörjningssystemet'' is a compound of the words for research information supply system.

The arbitrariness of segmenting units based on white space becomes especially clear when considering translations between languages. As one example, the concept of a ``knife sharpener'' is realized as two white-space delimited strings of characters in English, one in Swedish (``knivslip''), and three in Spanish (``afilador de cuchillo''). 

Segmentation is thus as non-trivial as it is foundational for NLP. Consequently, there exists a large body of work on segmentation algorithms (often driven by the need for segmenting languages other than English). Examples include \citet{Webster:1992,Chen:1992,saffran1996word,Beeferman1999,kiss:2006,Huang:2007}. Related areas (from the perspective of segmentation) such as multiword expressions and morphological normalization also have a rich literature of prior art. For multiword expressions, see e.g.~\citet{sag2002multiword,BaldwinK10,Constant:2017}, and for morphological normalization see e.g.~\citet{porter1980algorithm,kimmo,Yamashita:2000:LIM:974147.974179}.

In recent years, interest have begun to shift towards the use of {\em character-level} techniques, which bypass the problem of segmentation by simply operating on the raw character sequence. Much of this work is driven by research on deep learning, and techniques inspired by neural language models \citep{Sutskever:2011:GTR:3104482.3104610,Kim:2016:CNL:3016100.3016285}. In theory, such models can learn task-specific segmentations of the input that are optimal for solving whatever task the network is trained to perform.

The approach presented in this paper is inspired by character-level modeling, but in contrast to such techniques we seek a {\em task-independent} and {\em objective} segmentation of text. Our work is motivated by the idea that {\em if} there exists an optimal and language-invariant segmentation of text, it should be based on statistical properties of language rather than heuristics. We argue that such a segmentation exists, and introduce a novel type of data-driven segmented unit: the {\em recursion-gram} or \textit{r-gram} in short. The name is inspired by the n-gram introduced by \cite{shannon:1948}, who used it to explore language modeling in the context of information entropy, which was also introduced in the same paper. Our approach is inspired by information theoretic concerns.

In the applications where r-grams can be used, it replaces segmentation but not necessarily normalization. R-grams capture a range of semantic units from morphemes (or more generally, parts of words) to words to compounds to multi-word units, all based on simple frequency statistics. 
In this paper, we demonstrate an algorithm for computing one type of r-grams, and discuss novel observations and characteristics of the statistical distribution of natural language. We then demonstrate how r-grams can be used as basic building blocks in embeddings, and evaluate the resulting embeddings using both monolingual and multilingual test sets. We conclude the paper with some directions for future research.

\section{R-grams and compression algorithms}
Given a sequence over a finite alphabet, an r-gram is a variable length subsequence, derived by a set of well defined statistical rules, segmenting the original sequence into a set of subsequences.

\subsection{A first class of r-grams}
The fundamental idea of r-grams is deceptively simple. Given a sequence of discrete symbols sampled from a finite alphabet, find \textit{the most common pair} of adjacent symbols and \textit{replace} all instances of the pair with instances of a new single symbol, extending the alphabet by one, repeat until no more pairs can be found or some other criterion is fulfilled. 

\begin{table*}[t]
  \small 
  \begin{center}
  \begin{tabular}{llll}
  Iteration & Sequence & Alphabet & Replacement\\
  \hline
  0 & $s=<\beta,\beta,\beta,\alpha,\beta,\beta,\beta,\alpha,\beta,\beta,\beta>$ & $A={\alpha,\beta}$ & $<\beta, \beta> \rightarrow \gamma$ \\
  1 & $s=<\gamma,\beta,\alpha,\gamma,\beta,\alpha,\gamma,\beta>$ & $A={\alpha,\beta,\gamma}$ & $<\gamma, \beta> \rightarrow \delta$ \\
  2 & $s=<\delta,\alpha,\delta,\alpha,\delta>$ & $A={\alpha,\beta,\gamma,\delta}$ & \\
  \end{tabular}
  \end{center}
  \caption{Procedure to derive r-grams.}
  \label{tab:procedure}
\end{table*}

Table \ref{tab:procedure} illustrates an example where we have a sequence $S$ and an alphabet $A$. We show the first two iterations of the algorithm, at each step identifying the most common pair in the sequence and replacing it by a new symbol. Two new symbols $\gamma$ and $\delta$ are introduced. We observe a hierarchical structure where $\delta$ contains $\gamma$ which in turn contains symbols from the original alphabet. $\delta$ can thus be expanded into three elements from the original alphabet $\delta = (\beta, \beta, \beta)$.
	The observant reader might notice that there are some cases that require additional definitions. If two pairs overlap, as in the original sequence in the example, a rule for which pair to replace first has to be defined. In this example the rule was that the first from left to right observed pair is replaced. Another case is when there are more than one alternative for the most common pair, when the pairs has an equal amount of observations, then a rule on which pair to prioritize has to be defined. In the example above two r-grams were created: $\gamma = <\beta, \beta>$ and $\delta = <\beta, \beta, \beta>$.

If \textit{n} iterations of this procedure are performed on a sequence, the sequence is compressed, but the alphabet is expanded. Given that the compression of the sequence is larger than the expansion of the alphabet, we end up with a more compact representation of the underlying sequence. This exact procedure turns out to be an excellent compression algorithm named re-pair in the family of dictionary-based compression \citep{larsson2000off}. A remarkable property of this procedure is that, if the sequence is generated by an ergodic process, the segmented sequence becomes asymptotically Markov as the procedure is continually applied \citep{benedetto2006non}. 

A close relative to the re-pair algorithm is Byte Pair Encoding (BPE) \citep{gage1994bpe}, first used in the context of segmentation  by \cite{schuster2012japanese} and recently popularized in within deep learning by \cite{sennrich2015neural}. The segmentation method has primarily been used for finding subword units for later processing in recurrent neural networks, e.g. \cite{wu2016google} \cite{sennrich2015neural}. 

Published libraries for Byte Pair Encoding as segmentation exists in the form of, e.g. SentencePiece \cite{kudo2018sentencepiece}, and the resulting segments are commonly referred to as either ``sentencepieces'' or ``wordpieces'', the latter stressing their use as \emph{sub}word units. Functionally, the difference between such segmentation procedures and the r-gram algorithm is small, if at all existent. Crucially, however, we are interested in the \emph{properties} of the segmented units (which we call r-grams) and the grammar they form, rather than their use as a preprocessing step. 

\subsection{Implementation details}

The naive r-gram algorithm runs in quadratic time relative to the sequence length: find the most common pair in linear time, merge it, and repeat the process. This is prohibitively expensive. Thankfully there exists algorithms (namely re-pair and BPE) that recalculates the pair-frequencies in an efficient way, resulting in linear time algorithms. We have implemented a slightly modified version of the re-pair algorithm laid out in \cite{larsson2000off} that allows for other stopping criteria and accounts for document and sentence boundaries:

\noindent
\textbf{Stopping criterion.} We define two stopping criteria for the merges of the r-grams, which we simply call minimum frequency and maximum vocabulary. The minimum frequency criterion states that a new r-gram can be merged if its frequency exceeds the minimum frequency threshold, and the maximum vocabulary criterion simple states that new r-grams can be merged as long as the size of the vocabulary does not exceed the maximum vocabulary threshold. 

\noindent
\textbf{Sequences boundaries.} In natural language there are segmentations that signal a new local context such as sentence, paragraph or document boundaries. We generalize our statistics and alphabet collection over these boundaries but we do not create r-grams that overlap them.

The result of the re-pair compression algorithm on sequence $S$ is (\textbf{1}) a mapping from r-grams to their constituent parts (e.g. $\gamma \rightarrow \langle \beta, \beta \rangle$ from example \ref{tab:procedure}) and (\textbf{2}) a compressed sequence $S_c$ of the original sequence $S$, where $S_c$ is a sequence of r-grams rather than symbols from the original alphabet. By applying the mapping recursively down to the terminal symbols, the original sequence can be restored. When using r-grams as a segmentation technique, the sequence $S_c$ is taken to be a segmentation of $S$.

\section{Frequency distribution of data}

\begin{figure*}[t]
  \centering
  \includegraphics[width=\textwidth]{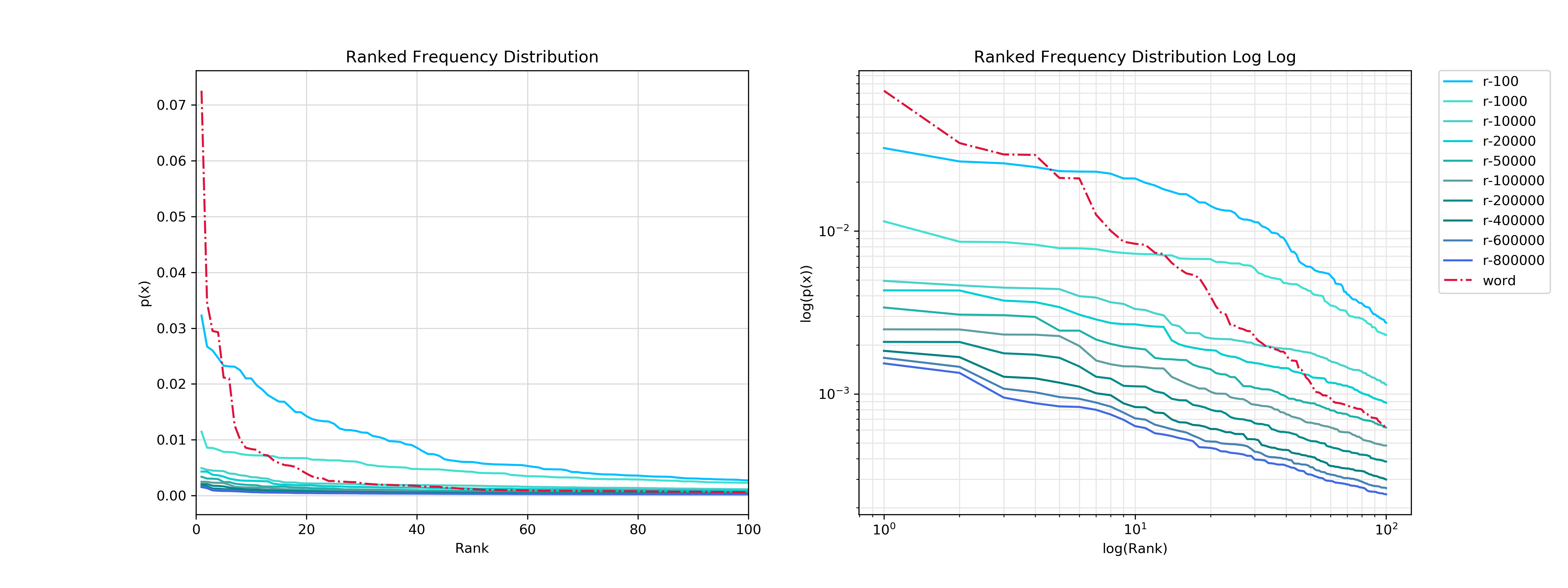}
  \caption{Ranked word and r-gram frequency distribution for the first hundred items in a subset of English Wikipedia.}
  \label{fig:frequency}
\end{figure*}

It is a well-known fact that the vocabulary of natural languages as segmented by traditional approaches follow a Zipfian distribution \citep{zipf32selective}. It is also a well-known fact that the majority of the frequency spectrum of traditionally segmented natural language is comprised of a small number of very high-frequent items, which are normally referred to as {\em stop words}. These high-frequent items are normally viewed as semantically vacuous, and are therefore generally not included in NLP applications. This practice has been around since the 1950s, when Hans Peter Luhn connected the ``resolving power'' of words in language to their frequency distribution \citep{luhn1958automatic}. Current methods in NLP still use basically the same type of algorithmic compensation for the power law distribution of word tokens in written text, whether it is the use of inverse document frequency in document processing applications, or subsampling \citep{mikolov2013distributed}, mutual information \citep{Church:1990:WAN:89086.89095}, or incremental frequency weighting \citep{Sahlgren:2016} in word embeddings. There is even debate whether the Zipfian distribution is an inherent language-specific feature or an emergent phenomenon sprung from the process of drawing and counting various-length character sequences from a finite alphabet \citep{piantadosi2014zipf}. 

From an information theoretic and information entropy perspective, a uniform distribution carries the most surprise \citep{shannon:1948} and thus also the most information. The Zipfian distribution belongs to the power law family and is highly non-uniform. Keeping both the practice of throwing away stop words and the information entropy perspective in mind, there is reason to believe that there exists more informative segmentations than word level segmentation for natural language.

Figure \ref{fig:frequency} shows the ranked frequency distributions for the 100 most high-frequent words and r-grams in a subset of English Wikipedia. The r-grams are computed over an increasing number of iterations (the $r$ parameter), and the figure clearly shows how the frequency distribution is flattened as the number of iterations of the r-gram algorithm is increased. This is a natural consequence of the algorithm, since it finds common elements and reforges some of them into new elements, reducing the frequency of common elements. In of itself this observation does not hold much value, but inspecting the r-grams created it seems that they capture semantic regularities such as morphemes, words and multi word units. Table \ref{tab:examples} demonstrates the effect of applying the algorithm to a 735MB sample text drawn from English Wikipedia. 
Note that after 100 iterations, the algorithm has formed the copula (``is'') and a determiner (``a''). After 1000 iterations, it has collapsed these into a common unit (``is a'') as well as the word ``born'' and the beginning of the collocation ``united states''. After 400 000 iterations, the algorithm has learned several long sequences, such as ``is a former'' and ``united states ambassador to''.

\begin{table*}
\resizebox{\textwidth}{!}{
\texttt{
\begin{tabular}{l l}
\toprule
Merges & Text\\
\hline
10    &r a n n e b er g er \_ ( b o r n \_ 1 9 4 9 ) \_ i s \_a \_ f o r m er \_ u\\
100   &r an n e b er g er\_ ( b or n \_ 19 4 9 ) \_ is \_a\_ for m er\_ un it ed\_ \\
1000  &ran ne ber g er\_ ( born \_ 194 9 ) \_is\_a\_ form er\_ united\_stat es \_am\\
20000 &ran ne berg er\_( born\_ 1949 )\_is\_a\_former\_ united\_states \_ambassador \\
100000&ran ne berg er\_(born\_ 1949 )\_is\_a\_former\_ united\_states \_ambassador\_t\\
400000&ran neberg er\_(born\_ 1949 )\_is\_a\_former\_ united\_states\_ambassador\_to\_\\
\bottomrule
\end{tabular}
}}
\caption{Textual example of r-grams being merged from English Wikipedia}
\label{tab:examples}
\end{table*}

Note that this has been learned from the statistics of the sequence alone, with all characters being treated as equal with no specific rules for whitespace or other special characters, with the exception for sequence separators such as newline. The important thing to note is that the r-gram algorithm learns units that would normally be discarded in NLP applications, since they contain (or, in the extreme case, consist entirely of) stop words. As an example, the phrase ``has yet to be'' constitutes a semantically useful unit that would be completely discarded when using standard stop word filtering. 

\section{Experiments} 

\subsection{R-grams in word embeddings}

The domain of NLP that focuses specifically on the {\em semantics} of units of language is called {\em distributional semantics}, where semantics is modeled using {\em distributional vectors} or {\em word embeddings}. Word embeddings encode semantic similarity by minimizing distance between vectors in a latent space, which is defined by co-occurrence information. Many methods for creating word embeddings have been proposed \citep{turney2010frequency}. Segmentation, as a preprocessing step, has a significant impact on the quality of word embeddings. The standard procedure is to simply rely on the white-space heuristic, and to remove all punctuation. This invariably leads to conflation of collocations in the distributional representations, and to problems with out of vocabulary items. 

To counter such problems, one may use preprocessing techniques to detect significant multiword expressions \citep{mikolov2013distributed} and morphological normalization \citep{Bullinaria:Levy:2012}, or one may try to incorporate string similarity into the distributional representation \citep{bojanowski2016enriching}, or detect collocations directly from the vector properties \citep{Sahlgren:2016}. 

A radically different approach, suggested by \cite{oshikiri2017segmentation}, is to produce embeddings for a subset of all possible character n-grams. This alleviates the need for preprocessing completely, but requires delimiting the subset with respect to the size of the n-grams, and their frequency of occurrence. \cite{hinken} also operates of character n-grams, but uses a random segmentation of the data. R-grams is similar in spirit to these previous approaches, but in contrast to the parameters required by \cite{oshikiri2017segmentation}, r-grams put no restrictions on the size of the units, or on their frequencies (except for the minimum frequency stopping criterion).

In order to demonstrate the applicability of r-grams for building word embeddings, we use 
a 735MB\footnote{The quality of the r-grams seem to correlate strongly to the amount of data they are derived from, more data equals better semantic representations. Our selected data size was dependent on the available RAM on the machine used for experiments.} subset of English Wikipedia for this experiment. The only preprocessing used before creating r-grams is lowercasing, for embeddings we also substitute numbers $0-9$ with $N$ and remove leading and trailing whitespaces from the r-grams. When building embeddings, we use skipgram with subword units \citep{bojanowski2016enriching}, a window size of $2$, and evaluate the models on standard \textit{single word} English embedding benchmarks\footnote{\url{https://github.com/kudkudak/word-embeddings-benchmarks}}. It is worth noting that the skipgram model uses subsampling 
of common words, which is an optimization introduced to compensate for the power law distribution in common vocabularies. Also, the skipgram model controls for collocations by dampening the impact of frequent collocations. This implies that the skipgram model might not be the optimal choice for creating embeddings from data driven segmentation. It was, however, the best performing model of those we tried during initial testing.

\begin{table}[!t]
\footnotesize 
\begin{center}
\begin{tabular}{ l l l}
    \toprule
    Test        	&	r-grams	&	words \\ [0.5ex] 
    \hline
    AP				&	0.58 			&	0.56 \\
    BLESS 			&	0.59 			&	0.75 \\
    Battig			& 	0.36 			&	0.40 \\
    ESSLI\_1a		&	0.73 			&	0.75 \\
    ESSLI\_2b		&	0.77 			&	0.80 \\
    ESSLI\_2c		&	0.62 			&	0.71 \\
    MEN				&	0.68 			&	0.73 \\
    MTurk			&	0.64 			&	0.67 \\
    RG65			&	0.66 			&	0.73 \\
    \bottomrule
    \end{tabular}
    \begin{tabular}{l l l}
    \toprule
    Test        	&	r-grams	&	words \\ [0.5ex] 
    \hline
    RW				& 	0.31 			&	0.39 \\
    SimLex999		&	0.36 			&	0.39 \\
    WS353			&	0.60 			&	0.67 \\
    WS353R			&	0.53 			&	0.61 \\
    WS353S			&	0.68 			&	0.70 \\
    Google			&	0.32 			&	0.33 \\
    MSR				&	0.39 			&	0.39 \\
    SemEval2012\_2	&	0.18 			&	0.21 \\
    && \\
    \bottomrule
    \end{tabular}
  \end{center}
  \caption{Comparison of word embeddings benchmarks using r-grams and words.}
  \label{tab:en_results}
  \end{table}

Table \ref{tab:en_results} shows the results of the embeddings produced using r-gram segmented data in comparison with whitespace segmented data. Note that the benchmark results in general are almost as good for the r-gram embedding as they are for the word embedding. In particular the analogy tests (Google and MSR) show no, or negligible, difference in the results between the r-gram embedding and the word embedding.
This is remarkable, since the r-grams have been learned directly from the character sequence, with no preconceptions of what constitutes viable semantic units. Taken by themselves, the scores for the r-gram embedding are competitive, and demonstrate the viability of the approach.

The benchmarks used in Table \ref{tab:en_results} only include single words. However, the r-grams range from parts of words to multiword expressions, strictly derived from the statistical distribution of the elements in the original sequence. In order to illustrate the qualitative properties of the r-gram embedding, Table \ref{tab:multiwords} show examples of the 10 nearest neighbors to a selected set of r-grams. Note that the r-grams may include punctuation as in ``dr. no'' and ``a hard day's'', and that the embedding includes phrases such as ``has yet to be'' (and all its neighbors) that would normally have been filtered out by stop word removal. The qualitative examples use the skipgram model without subword information.

\begin{table}[!t]
\footnotesize 
\begin{center}
\begin{tabular}{l l l}
  \toprule
  \# & {\bf 'back to the future'} & Cos \\ [0.5ex] 
  \hline
	1. 	& 'who framed roger rabbit' & 0.78 \\
	2. 	& 'dr. no'   				& 0.77 \\
	3. 	& 'show boat'        		& 0.76 \\
	4. 	& 'nightmare on elm street' & 0.75 \\
	5. 	& 'apocalypse now' 			& 0.75 \\
  \bottomrule
  \toprule
  \# & {\bf 'counterintelligence'} & Cos \\ [0.5ex] 
  \hline
	1. 	& 'counterterrorism'        	& 0.56 \\
	2. 	& 'intelligence community'  	& 0.55 \\
	3. 	& 'counter-terrorism'        	& 0.54 \\
	4. 	& 'intelligence'    			& 0.52 \\
	5. 	& 'advanced research project'   & 0.51 \\
  \bottomrule
  \end{tabular}
  \begin{tabular}{l l l}
  \toprule
  \# & {\bf 'has yet to be'} & Cos \\ [0.5ex] 
  \hline
	1. 	& 'has not been'        & 0.69 \\
	2. 	& 'has not yet been'   	& 0.68 \\
	3. 	& 'was never'        	& 0.59 \\
	4. 	& 'had not been'    	& 0.59 \\
	5. 	& 'has never been' 		& 0.59 \\
  \bottomrule
  \toprule
  \# & \textbf{'psychology'} & Cos \\ [0.5ex] 
  \hline
	1. 	& 'sociology'           & 0.69 \\
    2. 	& 'social psychology'   & 0.66 \\
	3. 	& 'anthropology'        & 0.65 \\
	4. 	& 'political theory'    & 0.64 \\
	5. 	& 'political science'   & 0.62 \\
  \bottomrule
\end{tabular}
\end{center}
\caption{Examples of the 5 nearest neighbors to four different targets in the r-gram embedding.}
\label{tab:multiwords}
\end{table}

\subsection{R-grams as a language agnostic segmentation technique}

To test whether or not r-gram segmentation is a viable language-agnostic segmentation technique we evaluate r-gram embeddings on the analogy test sets in \citep{grave2018learning}. These consist of (unbalanced) analogy tests for Czech, German, English, Spanish, Finnish, French, Hindi, Italian, Polish, Portuguese, and Chinese. For each language, we use a 750MB sample of Wikipedia, r-gram segmented with a stopping criteria of either \textit{minimum frequency} of $4$ or \textit{maximum vocabulary} of $800000$. As an additional preprocessing step we remove whitespace characters from the ends of r-grams: $\text{'example\_'} \rightarrow \text{'example'}$\footnote{This step --- while not strictly necessary --- was performed to better match the terms in the analogy tests.} The resulting, slightly modified, r-gram segmentation is then used to train r-gram embeddings using the skipgram model with subword units, as described in the previous subsection.

Despite the large variation across languages, the results in Table  \ref{tab:language_results} demonstrate that r-gram segmentation does indeed constitute a viable language-agnostic segmentation technique, albeit with poorer performance in the analogy tasks compared to regular segmentation.

\begin{table*}[ht]
\footnotesize 
  \centering
  \begin{tabular}{l l c c c c c c c c c c r} 
   \toprule
     & & CS & DE & ES & FI & FR & HI & IT & PL & PT & ZH & Average \\ [0.5ex] 
   \hline
   \multirow{2}{*}{Score} & r-gram	& 0.60 & 0.25 & 0.35 & 0.09	& 0.15 & 0.10 & 0.36 & 0.24 & 0.13 & 0.30 & 0.26 \\ 
   \cline{2-13}
   & baseline & 0.63 & 0.61 & 0.57 & 0.36	& 0.64 & 0.11 & 0.56 & 0.53 & 0.54 & 0.60 & 0.51 \\ 
   \hline
   \hline
    \multirow{2}{*}{Coverage} & r-gram & 0.66 & 0.54 & 0.64 & 0.85 & 0.67 & 0.40 & 0.52 & 0.38 & 0.61 & 0.96 & 0.62 \\
   \cline{2-13}
   & baseline & 0.77 & 0.79 & 0.94 & 0.95 & 0.88 & 0.71 & 0.81 & 0.70 & 0.79 & 1.00 & 0.83 \\
   \bottomrule
  \end{tabular}
  \caption{R-gram and baseline performance and coverage on the word analogy tasks. The baseline is taken from \cite{grave2018learning}}
  \label{tab:language_results}
\end{table*}

Part of the explanation for the relatively poor performance both here and the tests in the previous section is that the r-gram segmentation technique construct many near synonymous tokens. Table \ref{tab:pound_analogy} shows an example of this for the analogy query ``Great Britain is to the United States as Pound is to {\bf ?} `` in Finnish. The correct term according to the evaluation data set is `dollari`, which is not in the top ten candidates. However, `yhdysvaltain dollaria` (U.S. Dollar), is the second candidate. Dually, the top candidate is 'punt', which is a subword unit of 'punta', 'puntaa', 'puntin' et.c. We believe both of these types of near synonymous words, and their relative abundance in the r-gram vocabulary, has a detrimental effect on the word-based evaluation benchmarks. 
\begin{table}[!t]
\footnotesize 
\begin{center}
\begin{tabular}{l l l r}
  \toprule
    \# & $\textbf{'punta'} - \textbf{'englanti'} + \textbf{'yhdysvallat'}$ & Cos & Translation \\ [0.5ex] 
    \hline
    1. & 'punt' & 0.53 & 'Pound' \\
    2. & 'yhdysvaltain dollaria' & 0.50 & 'U.S. Dollar' \\ 
    3. & 'kun yhdysvallat' & 0.49 & 'When United States'\\
    4. & 'yhdysvaltain dollari' & 0.48 & 'U.S. Dollar'\\ 
    5. & 'yhdysvaltain dollarin' & 0.47 & 'U.S. Dollar'\\
  \bottomrule
\end{tabular}
\end{center}
\caption{Table showing analogy query candidates for finnish. The correct term according to the evaluation data set is 'dollari' which is only found as part of larger r-grams in the returned candidates.}
\label{tab:pound_analogy}
\end{table}

Going into a more qualitative view of what is represented by the r-gram embeddings in different languages, Table \ref{tab:multi} shows the nearest neighbors to two different acronyms (``vw'' and ``kgb'') in 6 different languages. The column marked \# indicates rank of the neighbor (i.e.~1 means the closest neighbor, and 7 means the seventh neighbor). The examples in Table \ref{tab:multi} demonstrate not only that the r-gram segmentation produces useful semantic units in all languages used in these experiments, but also that they constitute viable data for building embeddings; associated r-grams to ``vw'' are terms such as ``volkswagen'' and other automobile-related multiword units. The same applies to the neighbors of ``kgb''; neighbors are terms related to the secret police and security services. Again, note that all these terms were found by the unsupervised r-gram process.

\begin{table}[!t]
\small 
\begin{center}
\begin{tabular}{l l l}
  \toprule
  Lang. & \# & \textbf{'vw'} \\ [0.5ex] 
  \hline
  Spanish & 1. & 'volkswagen passat'\\
  German & 1. & 'volkswagen'\\
  Czech & 1. & 'koncernu volkswagen'\\
  Finnish & 1. & 'volkswagen golf'\\
  French & 7. & 'volkswagen'\\
  Polish & 1. & 'volkswagen'\\
  \bottomrule
  \end{tabular}
  \begin{tabular}{l l l}
  \toprule
  Lang. & \# & \textbf{'kgb'} \\ [0.5ex] 
  \hline
  Spanish & 4. & 'polic\'{i}a secreta' \\
  German & 1. & 'geheimdienstes' \\
  Czech & 2. & 'st\'{a}tn\'{i} bezpe\v{c}nosti' \\
  Finnish & 8. & 'yhdysvaltain \\
  & & keskustiedustelupalvelu' \\
  French & 8. & 'service de renseignement' \\
  Polish & 5. & 'g\l\'{o}wnego zarz\k{a}du \\
  & & bezpiecze\'{n}stwa pa\'{n}stwowego' \\
  \bottomrule
\end{tabular}
\end{center}
\caption{Examples of nearest neighbors in the r-gram embeddings in different languages to two different acronyms.}
\label{tab:multi}
\end{table}

The examples in Table \ref{tab:multi} where chosen with the intent to highlight how short r-grams can be viewed as semantically similar neighbors to longer r-grams. Next we turn to a demonstration of how the r-gram embeddings can be mapped across languages using a recently proposed unsupervised projection model (MUSE) \citep{lample2017unsupervised}. Their method leverages adversarial training to learn a linear mapping from a source to a target space, aligning embeddings trained on separate data allowing us to translate by finding similar vectors between the embeddings. Table \ref{tab:muse} demonstrates examples of translation between German and Spanish. In the first case we see how a single word in German (``k\"{u}rzer'', eng. ``shorter'') is mapped to relevant multiword units in Spanish. Note that the only difference between the first and second Spanish neighbor is the comma at the end. In the second example we see how a multiword unit in Spanish (``las ideas'', eng. ``the ideas'') is mapped to relevant single word units in German.

\begin{table}
\small 
\begin{center}
\begin{tabular}{l l}
  \toprule
  \multicolumn{2}{c}{German to Spanish} \\
  \hline
  \textbf{'kürzer'} & ('shorter')\\ [0.5ex]
  \hline
  'más corto' & ('shorter')\\
  'más corto,' & ('shorter') \\
  'mucho más larg' & ('much more larg(e)') \\
  'muy corto' & ('very short')\\
  'más cortos' & ('shorter') \\
  \bottomrule
  \end{tabular}
  \begin{tabular}{l l}
  \toprule
    \multicolumn{2}{c}{Spanish to German} \\
    \hline
  \textbf{'las ideas'}& ('the ideas')\\ [0.5ex]
  \hline
  'überzeugungen' & ('convictions')\\
  'tendenzen' & ('trends') \\
  'gedankengänge' & ('thought processes') \\
  'ideologien' & ('ideologies') \\
  'moralvorstellungen' & ('moral values')\\
  \bottomrule
\end{tabular}
\end{center}
\caption{Examples of crosslingual nearest neighbors using r-gram embeddings mapped with the MUSE algorithm. Words in parenthesis are English translation for the benefit of the reader.}
\label{tab:muse}
\end{table}

\section{Conclusions}

The main contribution of this paper is its novel perspective on segmentation as a statistical process operating on the raw character sequence. We believe that the application of this general process is not limited to language, but that it is generally applicable to compressible sequences of categorical data in order to find units and hierarchies. The fact that an r-gram is generated from a global context compression algorithm, and is also interpretable, is an interesting observation from the perspective of viewing AI as a compression problem \citep{mahoney1999text,legg2007universal}, which also suggests interesting directions for future work.

The substitution of the most common pair of types with a new type could be thought of as forming rules in a grammar. A lot of work has been done on inferring the smallest possible grammar (which turns out to be an NP complete problem \citep{charikar2005smallest}), as well as efficient grammar construction from local contexts \citep{nevill1997identifying}. The r-gram grammar (or graph) constitutes a very different type of grammar that contains both context, frequent collocations and natural subword units. It would be interesting to further investigate potential applications of this grammar; one interesting question is how the grammars differ between languages and in what ways they can be exploited in translation tasks, another very interesting possibility is to build embeddings directly on the grammar, since it records all necessary contextual information. Preliminary work indicates that generating embeddings directly from the r-gram grammar is a promising path going forward. 

\bibliographystyle{acl_natbib_nourl}
\bibliography{references}

\begin{thebibliography}{41}
\expandafter\ifx\csname natexlab\endcsname\relax\def\natexlab#1{#1}\fi

\bibitem[{Badino(2004)}]{Badino2004ChineseTW}
Leonardo Badino. 2004.
\newblock Chinese text word-segmentation considering semantic links among
  sentences.
\newblock In \emph{Proceedings of Interspeech}.

\bibitem[{Baldwin and Kim(2010)}]{BaldwinK10}
Timothy Baldwin and Su~Nam Kim. 2010.
\newblock Multiword expressions.
\newblock In \emph{Handbook of Natural Language Processing}, pages 267--292.
  Chapman and Hall/CRC.

\bibitem[{Beeferman et~al.(1999)Beeferman, Berger, and
  Lafferty}]{Beeferman1999}
Doug Beeferman, Adam Berger, and John Lafferty. 1999.
\newblock Statistical models for text segmentation.
\newblock \emph{Machine Learning}, 34(1):177--210.

\bibitem[{Benedetto et~al.(2006)Benedetto, Caglioti, and
  Gabrielli}]{benedetto2006non}
Dario Benedetto, Emanuele Caglioti, and Davide Gabrielli. 2006.
\newblock Non-sequential recursive pair substitution: some rigorous results.
\newblock \emph{Journal of Statistical Mechanics: Theory and Experiment},
  2006(09):P09011.

\bibitem[{Bojanowski et~al.(2017)Bojanowski, Grave, Joulin, and
  Mikolov}]{bojanowski2016enriching}
Piotr Bojanowski, Edouard Grave, Armand Joulin, and Tomas Mikolov. 2017.
\newblock Enriching word vectors with subword information.
\newblock \emph{Transactions of the Association for Computational Linguistics},
  5:135--146.

\bibitem[{Bullinaria and Levy(2012)}]{Bullinaria:Levy:2012}
John Bullinaria and Joseph~P. Levy. 2012.
\newblock Extracting semantic representations from word co-occurrence
  statistics: stop-lists, stemming, and svd.
\newblock \emph{Behavior Research Methods}, 44:890--907.

\bibitem[{Charikar et~al.(2005)Charikar, Lehman, Liu, Panigrahy, Prabhakaran,
  Sahai, and Shelat}]{charikar2005smallest}
Moses Charikar, Eric Lehman, Ding Liu, Rina Panigrahy, Manoj Prabhakaran, Amit
  Sahai, and Abhi Shelat. 2005.
\newblock The smallest grammar problem.
\newblock \emph{IEEE Transactions on Information Theory}, 51(7):2554--2576.

\bibitem[{Chen and Liu(1992)}]{Chen:1992}
Keh-Jiann Chen and Shing-Huan Liu. 1992.
\newblock Word identification for mandarin chinese sentences.
\newblock In \emph{Proceedings of the COLING}, pages 101--107.

\bibitem[{Church and Hanks(1990)}]{Church:1990:WAN:89086.89095}
Kenneth~Ward Church and Patrick Hanks. 1990.
\newblock Word association norms, mutual information, and lexicography.
\newblock \emph{Computational Linguistics}, 16(1):22--29.

\bibitem[{Constant et~al.(2017)Constant, Eryiğit, Monti, van~der Plas,
  Ramisch, Rosner, and Todirascu}]{Constant:2017}
Mathieu Constant, Gülşen Eryiğit, Johanna Monti, Lonneke van~der Plas,
  Carlos Ramisch, Michael Rosner, and Amalia Todirascu. 2017.
\newblock Multiword expression processing: A survey.
\newblock \emph{Computational Linguistics}, 43(4):837--892.

\bibitem[{Gage(1994)}]{gage1994bpe}
Philip Gage. 1994.
\newblock A new algorithm for data compression.
\newblock \emph{C Users J.}, 12(2):23–38.

\bibitem[{Grave et~al.(2018)Grave, Bojanowski, Gupta, Joulin, and
  Mikolov}]{grave2018learning}
Edouard Grave, Piotr Bojanowski, Prakhar Gupta, Armand Joulin, and Tomas
  Mikolov. 2018.
\newblock Learning word vectors for 157 languages.
\newblock In \emph{Proceedings of LREC}.

\bibitem[{Haspelmath(2011)}]{haspelmath:2011}
M.~Haspelmath. 2011.
\newblock The indeterminacy of word segmentation and the nature of morphology
  and syntax.
\newblock \emph{Folia Linguistica}, 45(1):31--80.

\bibitem[{Huang et~al.(2007)Huang, \v{S}imon, Hsieh, and
  Pr{\'e}vot}]{Huang:2007}
Chu-Ren Huang, Petr \v{S}imon, Shu-Kai Hsieh, and Laurent Pr{\'e}vot. 2007.
\newblock Rethinking chinese word segmentation: Tokenization, character
  classification, or wordbreak identification.
\newblock In \emph{Proceedings of ACL}, pages 69--72.

\bibitem[{Kim et~al.(2016)Kim, Jernite, Sontag, and
  Rush}]{Kim:2016:CNL:3016100.3016285}
Yoon Kim, Yacine Jernite, David Sontag, and Alexander~M. Rush. 2016.
\newblock Character-aware neural language models.
\newblock In \emph{Proceedings of AAAI}, pages 2741--2749.

\bibitem[{Kiss and Strunk(2006)}]{kiss:2006}
Tibor Kiss and Jan Strunk. 2006.
\newblock Unsupervised multilingual sentence boundary detection.
\newblock \emph{Computational Linguistics}, 32(4):485--525.

\bibitem[{Koskenniemi(1996)}]{kimmo}
Kimmo Koskenniemi. 1996.
\newblock Finite state morphology in information retrieval.
\newblock \emph{Natural Language Engineering}, 2.

\bibitem[{Kudo and Richardson(2018)}]{kudo2018sentencepiece}
Taku Kudo and John Richardson. 2018.
\newblock Sentencepiece: A simple and language independent subword tokenizer
  and detokenizer for neural text processing.
\newblock In \emph{Proceedings of the 2018 Conference on Empirical Methods in
  Natural Language Processing: System Demonstrations}, pages 66--71.

\bibitem[{Lample et~al.(2018)Lample, Conneau, Denoyer, and
  Ranzato}]{lample2017unsupervised}
Guillaume Lample, Alexis Conneau, Ludovic Denoyer, and Marc'Aurelio Ranzato.
  2018.
\newblock Unsupervised machine translation using monolingual corpora only.
\newblock In \emph{Proceedings of ICLR}.

\bibitem[{Larsson and Moffat(2000)}]{larsson2000off}
N~Jesper Larsson and Alistair Moffat. 2000.
\newblock Off-line dictionary-based compression.
\newblock \emph{Proceedings of the IEEE}, 88(11):1722--1732.

\bibitem[{Legg and Hutter(2007)}]{legg2007universal}
Shane Legg and Marcus Hutter. 2007.
\newblock Universal intelligence: A definition of machine intelligence.
\newblock \emph{Minds and Machines}, 17(4):391--444.

\bibitem[{Luhn(1958)}]{luhn1958automatic}
Hans~Peter Luhn. 1958.
\newblock The automatic creation of literature abstracts.
\newblock \emph{IBM Journal of research and development}, 2(2):159--165.

\bibitem[{Mahoney(1999)}]{mahoney1999text}
Matthew~V Mahoney. 1999.
\newblock Text compression as a test for artificial intelligence.
\newblock In \emph{AAAI/IAAI}, page 970.

\bibitem[{Mikolov et~al.(2013)Mikolov, Sutskever, Chen, Corrado, and
  Dean}]{mikolov2013distributed}
Tomas Mikolov, Ilya Sutskever, Kai Chen, Greg~S Corrado, and Jeff Dean. 2013.
\newblock Distributed representations of words and phrases and their
  compositionality.
\newblock In \emph{Proceedings of NIPS}, pages 3111--3119.

\bibitem[{Nevill-Manning and Witten(1997)}]{nevill1997identifying}
Craig~G Nevill-Manning and Ian~H Witten. 1997.
\newblock Identifying hierarchical structure in sequences: A linear-time
  algorithm.
\newblock \emph{Journal of Artificial Intelligence Research}, 7:67--82.

\bibitem[{Oshikiri(2017)}]{oshikiri2017segmentation}
Takamasa Oshikiri. 2017.
\newblock Segmentation-free word embedding for unsegmented languages.
\newblock In \emph{Proceedings of EMNLP}, pages 767--772.

\bibitem[{Piantadosi(2014)}]{piantadosi2014zipf}
Steven~T Piantadosi. 2014.
\newblock Zipf’s word frequency law in natural language: A critical review
  and future directions.
\newblock \emph{Psychonomic bulletin \& review}, 21(5):1112--1130.

\bibitem[{Porter(1980)}]{porter1980algorithm}
Martin~F Porter. 1980.
\newblock An algorithm for suffix stripping.
\newblock \emph{Program}, 14(3):130--137.

\bibitem[{Saffran et~al.(1996)Saffran, Newport, and Aslin}]{saffran1996word}
Jenny~R Saffran, Elissa~L Newport, and Richard~N Aslin. 1996.
\newblock Word segmentation: The role of distributional cues.
\newblock \emph{Journal of memory and language}, 35(4):606--621.

\bibitem[{Sag et~al.(2002)Sag, Baldwin, Bond, Copestake, and
  Flickinger}]{sag2002multiword}
Ivan~A Sag, Timothy Baldwin, Francis Bond, Ann Copestake, and Dan Flickinger.
  2002.
\newblock Multiword expressions: A pain in the neck for nlp.
\newblock In \emph{Proceedings of CICLing}, pages 1--15.

\bibitem[{Sahlgren et~al.(2016)Sahlgren, Gyllensten, Espinoza, Hamfors, Holst,
  Karlgren, Olsson, Persson, and Viswanathan}]{Sahlgren:2016}
Magnus Sahlgren, Amaru~Cuba Gyllensten, Fredrik Espinoza, Ola Hamfors, Anders
  Holst, Jussi Karlgren, Fredrik Olsson, Per Persson, and Akshay Viswanathan.
  2016.
\newblock The {G}avagai {L}iving {L}exicon.
\newblock In \emph{Proceedings of LREC}.

\bibitem[{Schuster and Nakajima(2012)}]{schuster2012japanese}
Mike Schuster and Kaisuke Nakajima. 2012.
\newblock Japanese and korean voice search.
\newblock In \emph{ICASSP}, pages 5149--5152.

\bibitem[{Sch{\"u}tze(2017)}]{hinken}
Hinrich Sch{\"u}tze. 2017.
\newblock Nonsymbolic text representation.
\newblock In \emph{Proceedings of EACL}, pages 785--796.

\bibitem[{Sennrich et~al.(2016)Sennrich, Haddow, and
  Birch}]{sennrich2015neural}
Rico Sennrich, Barry Haddow, and Alexandra Birch. 2016.
\newblock Neural machine translation of rare words with subword units.
\newblock In \emph{Proceedings of ACL}, pages 1715--1725.

\bibitem[{Shannon(1948)}]{shannon:1948}
Claude~E. Shannon. 1948.
\newblock A mathematical theory of communication.
\newblock \emph{Bell System Technical Journal}, 27(3):379--423.

\bibitem[{Sutskever et~al.(2011)Sutskever, Martens, and
  Hinton}]{Sutskever:2011:GTR:3104482.3104610}
Ilya Sutskever, James Martens, and Geoffrey Hinton. 2011.
\newblock Generating text with recurrent neural networks.
\newblock In \emph{Proceedings of ICML}, pages 1017--1024.

\bibitem[{Turney and Pantel(2010)}]{turney2010frequency}
Peter~D Turney and Patrick Pantel. 2010.
\newblock From frequency to meaning: Vector space models of semantics.
\newblock \emph{Journal of artificial intelligence research}, 37:141--188.

\bibitem[{Webster and Kit(1992)}]{Webster:1992}
Jonathan~J. Webster and Chunyu Kit. 1992.
\newblock Tokenization as the initial phase in nlp.
\newblock In \emph{Proceedings of COLING}, pages 1106--1110.

\bibitem[{Wu et~al.(2016)Wu, Schuster, Chen, Le, Norouzi, Macherey, Krikun,
  Cao, Gao, Macherey et~al.}]{wu2016google}
Yonghui Wu, Mike Schuster, Zhifeng Chen, Quoc~V Le, Mohammad Norouzi, Wolfgang
  Macherey, Maxim Krikun, Yuan Cao, Qin Gao, Klaus Macherey, et~al. 2016.
\newblock Google's neural machine translation system: Bridging the gap between
  human and machine translation.
\newblock \emph{arXiv preprint arXiv:1609.08144}.

\bibitem[{Yamashita and Matsumoto(2000)}]{Yamashita:2000:LIM:974147.974179}
Tatsuo Yamashita and Yuji Matsumoto. 2000.
\newblock Language independent morphological analysis.
\newblock In \emph{Proceedings of ANLC}, pages 232--238.

\bibitem[{Zipf(1932)}]{zipf32selective}
G.~K. Zipf. 1932.
\newblock \emph{Selective Studies and the Principle of Relative Frequency in
  Language}.
\newblock Harvard University Press.

\end{thebibliography}

\end{document}